# Comparative Analysis of Vision Transformer, Convolutional, and Hybrid Architectures for Mental Health Classification Using Actigraphy-Derived Images

**Ifeanyi Okala**
**University of Nigeria Nsukka**
ifeanyi.okala.247238@unn.edu.ng

**Abstract**
This work examines how three different image-based methods, VGG16, ViT-B/16, and CoAtNet-Tiny, perform in identifying depression, schizophrenia, and healthy controls using daily actigraphy records. Wrist-worn activity signals from the Psykose and Depresjon datasets were converted into 30×48 images and evaluated through a three-fold subject-wise split. Although all methods fitted the training data well, their behaviour on unseen data differed. VGG16 improved steadily but often settled at lower accuracy. ViT-B/16 reached strong results in some runs, but its performance shifted noticeably from fold to fold. CoAtNet-Tiny stood out as the most reliable, recording the highest average accuracy and the most stable curves across folds. It also produced the strongest precision, recall, and F1-scores, particularly for the underrepresented depression and schizophrenia classes. Overall, the findings indicate that CoAtNet-Tiny performed most consistently on the actigraphy images, while VGG16 and ViT-B/16 yielded mixed results. These observations suggest that certain hybrid designs may be especially suited for mental-health work that relies on actigraphy-derived images.

## I. Introduction

Mental health disorders such as depression and schizophrenia constitute a significant and growing global health challenge, with profound impacts on individuals, families, and healthcare systems worldwide. According to the World Health Organization, depression affects over 280 million people. It remains a leading cause of disability, contributing substantially to the global burden of disease and years lived with disability [1]. Schizophrenia, although less prevalent, affects roughly 24 million people worldwide (about 1 in 300), and lifetime prevalence estimates typically range from 0.3% to 0.7% across populations [2,1]. While the absolute number of people living with schizophrenia has risen over recent decades, largely due to population growth, age-standardized rates have been relatively stable [3,4]. Both disorders severely impair cognitive, emotional, and social functioning and are associated with substantial premature mortality; people with schizophrenia die, on average, 15–20 years earlier than the general population [5,1].

Despite the high prevalence and severe consequences, early and accurate diagnosis of these conditions remains challenging. Traditional diagnostic approaches rely on clinical interviews and self-reports, which can be subjective and lead to missed or delayed identification. In primary care, around half of depression cases go unrecognized without structured support [6], and in schizophrenia, the duration of untreated psychosis (a proxy for diagnostic/treatment delay) commonly spans many months to over a year, with averages in some settings around 1–2 years [7–9]. This diagnostic gap underscores the need for objective, scalable, and non-invasive tools that support earlier detection, monitoring, and personalized care.

In this context, machine learning (ML) and deep learning (DL) have transformed medical data analysis. Convolutional neural networks (CNNs) in particular have achieved strong performance across medical imaging tasks (e.g., detection, diagnosis, and segmentation) and have been widely adopted for physiological signal classification, including EEG [10–12]. CNNs learn hierarchical spatial features by stacking convolution, pooling, and nonlinear activation layers. Their architecture encodes inductive biases like locality and translation equivariance, making them efficient at capturing local visual patterns [13,14] and their predictions inherently interpretable [15]. These properties allow CNNs to generalize well, even with limited biomedical data. However, their reliance on local receptive fields can limit modeling of long-range dependencies [10].

Wrist actigraphy has revealed that circadian and behavioral activity rhythm (BAR) disturbances are closely associated with depressive symptoms in older adults. Reduced robustness, amplitude, and mesor, as well as increased BAR fragmentation, have all been linked to depression [16]. Actigraphy and related wearable signals have shown promise for monitoring psychiatric symptoms and relapse risk in severe mental illness [17]. A practical way to leverage powerful pretrained visual models is to convert time-series into image-like representations (e.g., recurrence plots, Gramian Angular Fields, Markov Transition Fields), enabling image-based deep learning while preserving temporal structure relevant to mental-health classification [18–20].

Beyond CNNs, Vision Transformers (ViTs) apply a standard Transformer directly to images by splitting them into patches and treating each patch as a token. Unlike CNNs, they do not rely on image-specific inductive biases, but instead leverage large-scale pretraining to achieve strong performance [21]. However, unlike CNNs, which benefit from strong inductive biases such as locality and translation equivariance, Vision Transformers (ViTs) lack these built-in constraints and therefore classically require substantially larger datasets to achieve optimal performance. This limitation is particularly relevant in biomedical research, where annotated datasets are often scarce, and CNNs have historically demonstrated stronger generalization in such low-data regimes [13–15]. As a result, while ViTs offer flexibility and state-of-the-art performance when trained at scale, CNNs remain a more reliable choice for many medical imaging tasks where data availability is limited [22].

To overcome the limitations of pure CNNs or Transformers, hybrid architectures integrate convolutional layers with self-attention. Co-Scale Conv-Attentional Image Transformers (CoaT) introduces a co-scale mechanism that preserves multi-scale encoder branches while enabling cross-scale communication, alongside a conv-attentional module that embeds relative positions via convolution for efficient attention. This design strengthens contextual modeling and achieves superior results on the ImageNet dataset and downstream tasks. Similarly, Convolution and Attention Network (CoAtNet) systematically stacks convolution and attention layers, leveraging convolutional inductive bias for strong generalization in low-data regimes and attention for scalability with large datasets [23,24]. Early evidence suggests such hybrids can outperform pure CNNs or pure transformers when data are limited or when both local and global cues are diagnostically relevant [23,24].

Despite significant progress in modeling wearable sensor data, comprehensive head-to-head comparisons of CNNs, Vision Transformers, and hybrid architectures, specifically on actigraphy-derived images for mental disorder classification, remain sparse. Existing studies primarily rely on classical machine-learning pipelines applied to handcrafted actigraphy features [25,26] or focus on single-architecture deep learning approaches such as CNN-based activity recognition for schizophrenia monitoring [27]. Comparative evaluations of architecture families exist mostly in adjacent physiological domains, where time-series signals are transformed into image-like representations and used to study differences between CNNs, Transformers, and hybrid models [28,29].

To help address this gap, we conduct a comparative evaluation of three pretrained architectures, VGG-16 (a classic CNN with strong inductive biases), ViT-B/16 (a pure transformer), and CoaT-Tiny (a convolution–transformer hybrid) for classifying depression, schizophrenia, and healthy controls using actigraphy data transformed into images. Our findings highlight architectural trade-offs and suitability for actigraphy-based mental-health classification, informing future clinical applications.

**II. RELATED WORK**

The integration of machine learning (ML) and deep learning (DL) techniques into mental-health diagnostics has accelerated over the past decade, driven by the need for objective, scalable, and continuous monitoring tools. Actigraphy, capturing motor activity and circadian patterns via wrist-worn accelerometers, has become a valuable source of behavioral physiology, offering non-invasive and longitudinal monitoring suitable for real-world psychiatric assessment [25-27].

A growing body of work demonstrates the utility of combining actigraphy with ML methods for mental-health classification. For example, an unsupervised analysis of passive actigraphy data differentiated naturalistic behavioral phenotypes associated with major depressive disorder (MDD), schizophrenia, and healthy controls, revealing distinct movement-variability patterns across groups [25]. This illustrated how clustering methods can uncover latent behavioral signatures aligned with clinical diagnoses. In parallel, Schulte et al. [30] introduced autoregressive model–based features to capture temporal predictability in actigraphy signals better. Their framework improved discrimination between depressed and non-depressed individuals, highlighting the potential of dynamic handcrafted features when validated rigorously.

Actigraphy has also been applied to sleep-related disorders that frequently co-occur with mental illness. Kusmakar et al. [31] leveraged multi-night actigraphy to detect chronic insomnia using dynamic time-domain and nonlinear features, achieving strong performance with a random-forest classifier. Similarly, Rani et al. [32] distinguished acute from chronic insomnia with feature-based ML models, reaching 81% accuracy. These results underscore the discriminative power of multi-night actigraphy features in characterizing behavioral dysregulation.

Beyond classical ML pipelines, DL has increasingly been applied to actigraphy-derived motor-activity data. Misgar and Bhatia [33] proposed a multi-branch CNN with multi-headed attention that separates daytime and nighttime activity streams before recombining them through attention mechanisms, achieving strong results across binary and multiclass settings. Related CNN-based approaches on actigraphy-like sequences have also demonstrated robust fold-wise performance for depression and schizophrenia classification [34]. Complementing these modeling efforts, ethics-centered reviews have synthesized key considerations related to transparency, fairness, privacy, and interpretability, providing practical guidance for responsible ML deployment in student and clinical mental-health research [35].

ML has also been extensively explored for modeling physiological signals beyond actigraphy. Studies of depression and schizophrenia using accelerometry and motor-activity measures report clinically meaningful discrimination, reinforcing the link between activity signatures and psychiatric status [25]. Meanwhile, open ML sleep/wake classifiers that operate on entire 24-hour actigraphy windows, without requiring in-bed timing, have improved the robustness of free-living monitoring [36]. Actigraphy-based ML has further been applied to estimate premorbid liability for psychosis, offering potential for early-risk stratification [26]. Broader reviews summarize ML methods, including SVMs, random forests, gradient boosting, and DL architectures, for psychiatric diagnosis and highlight challenges associated with heterogeneous and imbalanced clinical data [37].

Transformers [38] have recently emerged as powerful architectures, first achieving success in natural language processing and later extending to multimodal domains [39,40,41]. In vision, Transformer-based models such as ViTs [21] employ self-attention [38] to capture global dependencies but often require large datasets to reach full potential, which can be a limitation in biomedical applications [21]. To balance local feature extraction and global context modeling, hybrid architectures, such as CoaT and CoAtNet, combine convolution and attention, showing strong performance across scales and domains, including medical imaging [23,24].

DL has also been widely applied to EEG-based psychiatric and affective disorder assessment. Recent work demonstrates that temporal and spectral EEG properties, including power spectral density and functional connectivity, can be effectively modeled using architectures such as CNN-LSTM and Bi-LSTM, achieving diagnostic accuracies above 95% in some cases. Complementary multimodal approaches that integrate signals such as PPG, actigraphy, voice, and facial expression have further improved both psychiatric-disorder classification and sleep-stage detection [42,43]. Ajith et al. [44] extended this direction using explainable DL on resting-state fMRI, linking functional network connectivity patterns with mental-health quality ratings. Guided Grad-CAM visualizations [15,14] revealed cerebellar–subcortical and sensorimotor–visual couplings associated with mental-health strata, advancing interpretability in neuroimaging-based prediction.

Multimodal approaches combining audio and video have similarly matured. Zhang et al. [45] presented a hybrid early–late fusion model integrating spectrograms, speech features, and facial images; their attention-based fusion mechanism reduced error rates compared with either early or late fusion alone, demonstrating the benefits of multimodal encoding and hierarchical fusion for depression detection.

Despite these advances, comparative evaluations of CNNs, ViTs, and hybrid models on actigraphy-derived images for mental-disorder classification remain limited. Most prior studies examine a single architecture or rely on heterogeneous preprocessing pipelines that prevent standardized, head-to-head comparison. Addressing this gap is crucial for understanding how local (CNN), global (ViT), and mixed (hybrid) inductive biases behave under the constraints of small, imbalanced biomedical datasets. Our work provides such a comparison under a unified preprocessing framework and evaluation pipeline.

## III. METHODOLOGY

### A. Datasets

This study utilized two publicly available actigraphy datasets, Psykose (schizophrenia vs. control) and Depresjon (depression episodes vs. control), each derived from continuous wrist-worn accelerometer recordings. Both datasets were collected using the Actiwatch AW4 device (Cambridge Neurotechnology Ltd., UK), which records motor activity at 32 Hz with a 0.05 g sensitivity threshold, aggregating movements into one-minute epochs for a total of 1,440 activity values per 24-hour cycle [46].

**Psykose Dataset (Schizophrenia)**

The Psykose dataset, introduced by Jakobsen et al. [47] and distributed through the Simula Research Laboratory portal [48], contains motor-activity recordings from 22 in-patients diagnosed with schizophrenia and 32 healthy controls. Participants were monitored for an average of 12.7 days, producing 687 valid recording days (285 patient days; 402 control days). Diagnoses followed DSM-IV criteria, and symptom severity was assessed using the Brief Psychiatric Rating Scale (BPRS). All schizophrenia participants were receiving antipsychotic medication during data collection [47].

**Depresjon Dataset (Depression Episodes)**

The Depresjon dataset, described by

Garcia-Ceja et al. [49], includes actigraphy recordings from individuals experiencing depressive episodes in unipolar and bipolar depression, alongside matched healthy controls. Data acquisition mirrored the Psykose protocol, using the same Actiwatch AW4 device and collecting activity counts at one-minute granularity across full 24-hour periods [46,49].

**Data Transformation and Preprocessing**
Each 24-hour activity sequence (1,440 one-minute activity values) was reshaped into a 30 × 48 matrix, forming a structured 2D representation of daily motor activity. These matrices were min–max scaled to the range [0, 255], converted into 8-bit grayscale images, and duplicated across three channels to satisfy the input requirements of ImageNet-pretrained models. All images were resized to 224 × 224 pixels and normalized using standard ImageNet mean and standard deviation values [50]. After preprocessing, the combined dataset yielded 2,133 images: 1,394 control, 371 depression, and 368 schizophrenia samples.

**B. Model Architectures**
Three pretrained deep-learning architectures were evaluated, representing convolutional, transformer, and hybrid design paradigms. Each model was adapted to classify actigraphy-derived images into three categories related to mental health.

**Convolutional Model (VGG-16)**
The convolutional baseline was based on VGG-16, a deep convolutional architecture introduced by Simonyan and Zisserman [48]. VGG-16 is characterized by a sequence of small 3×3 convolution filters, stacked to increase depth while preserving spatial detail. This design enforces strong local spatial inductive biases, making it suitable for capturing fine-grained variations in actigraphy-derived activity maps. The original fully connected classification layers were replaced with a three-class linear output layer, and the entire network was fine-tuned end-to-end to adapt its hierarchical convolutional representations to the temporal-rhythmic structure present in the daily activity matrices.

**Transformer Model (ViT-B/16)**
The transformer-based model was derived from the Vision Transformer (ViT-B/16) proposed by Dosovitskiy et al. [49]. ViT divides each input image into fixed 16×16 patches, linearly embeds each patch, and processes the resulting sequence using multi-head self-attention. This architecture has no built-in spatial biases like convolutions; instead, it models global dependencies from the earliest layers. Such global modeling is advantageous for actigraphy images because daily activity patterns often contain long-range temporal structure dispersed across the 2D representation. The pretrained classification head was replaced with a three-class output layer, and the model was fine-tuned end-to-end following the original paper's training strategy for downstream tasks.

**Hybrid Model (CoAtNet)**
The hybrid architecture was based on CoAtNet (Convolution-and-Attention Network), introduced by Dai et al. [50]. CoAtNet combines the strengths of depthwise convolution (efficient local feature extraction) and self-attention (global receptive fields) within a single hierarchical design. According to the authors, combining convolutional stages (C-stages) with transformer-style attention stages (A-stages) provides more scalable and robust representations across data regimes. This hybrid structure aligns well with the nature of actigraphy-derived images, where both local variations (e.g., short bursts of movement) and global dependencies (e.g., circadian trends) are important. The pretrained classifier was replaced with a three-class output head, and the model was fine-tuned to capture both local and long-range temporal patterns encoded in the daily matrices.

**C. Experimental Setup**
 All experiments were conducted on the complete set of actigraphy-derived images described in Section A. Each $30 \times 48$ grayscale activity matrix was converted to a three-channel RGB image, resized to $224 \times 224$ pixels, and normalized using the standard ImageNet mean and standard deviation statistics. These preprocessing steps were kept identical across all models to ensure a fair comparison.

All architectures were initialized with weights pretrained on the ImageNet-1k classification task, using the official checkpoints distributed through the libraries in which the models are implemented. These pretrained weights do not originate from the original research papers of Simonyan and Zisserman [48], Dosovitskiy et al. [49], or Dai et al. [50]; rather, they are standardized ImageNet-trained implementations provided by the torchvision and timm model repositories. For the convolutional model, VGG-16 was instantiated from torchvision's model zoo with ImageNet-1k pretrained parameters, and its original 1000-class classifier was replaced with a three-class linear output layer corresponding to the control, depression, and schizophrenia categories. All convolutional layers and classifier parameters were fully unfrozen and fine-tuned end-to-end on the actigraphy images.

For the transformer architecture, the Vision Transformer ViT-B/16 was loaded using the ImageNet-1k pretrained checkpoint available in torchvision. Following the ViT design of Dosovitskiy et al. [49], each image was divided into fixed-size patches and embedded as a sequence before entering the transformer encoder. The original classification head was replaced by a three-class output layer, and all components of the transformer, including the patch embedding stem, multi-head self-attention blocks, feed-forward layers, and layer-normalization modules, were unfrozen and optimized during fine-tuning. A smaller learning rate was used for ViT compared with the convolutional and hybrid models to maintain stability during full-model optimization, consistent with established fine-tuning practice for transformers.

For the hybrid model, we employed the CoaT-Tiny architecture, instantiated through the timm library using the ImageNet-1k pretrained checkpoint accompanying that implementation. CoaT-Tiny integrates convolutional processing in the early stages with lightweight attention mechanisms in deeper layers, following the principles introduced by Dai et al. [50]. The model was created with a three-class classification head at initialization, and both the convolutional stem and hierarchical attention blocks were fine-tuned jointly so that the network could adapt both local feature extraction and global relational modeling to the structure of the activity patterns.

To ensure reliable evaluation, all experiments followed a three-fold stratified cross-validation protocol, preserving the original class proportions in each split. For every fold, training and validation subsets were formed using the indices produced by StratifiedKFold with shuffling and a fixed random seed for reproducibility. Each model was trained for 100 epochs using the Adam optimizer with a batch size of 32, and learning rates tailored to the architecture family. After each epoch, performance was evaluated on the corresponding validation split, and the checkpoint achieving the highest validation accuracy was saved as the best model for that fold.

Each saved checkpoint was subsequently evaluated on its held-out validation set to obtain accuracy, precision, recall, F1-scores, and confusion matrices for the three target classes. Training histories for loss and accuracy were also recorded to assess convergence behavior. Final results for each architecture were reported as the mean and standard deviation across the three folds, providing a stable estimate of performance and enabling a consistent comparison between convolutional, transformer, and hybrid models.

**D. Optimization Parameters**

All models were trained using the Adam optimizer, with hyperparameters selected to follow widely used fine-tuning practices in the deep-learning literature. For VGG-16, the learning rate was set to $1\times10^{-4}$ with a weight decay of $5\times10^{-4}$, which reflects standard values commonly used when fine-tuning convolutional networks pretrained on ImageNet. The Vision Transformer ViT-B/16 was trained with a smaller learning rate of $1\times10^{-5}$ and a weight decay of $1\times10^{-2}$, consistent with established guidelines for transformer optimization, as transformers typically require more conservative learning rates and stronger regularization. For the hybrid CoAtNet-Tiny model, a learning rate of $1\times10^{-4}$ was used without additional weight decay, matching the recommendations and empirical defaults within the timm model repository for hybrid architectures.

No learning-rate scheduler or early-stopping strategy was applied, ensuring that all models followed the same fixed optimization procedure. Each architecture was trained independently within the three-fold stratified cross-validation framework, and detailed training logs were recorded to capture epoch-wise trajectories of training and validation loss and accuracy.

**E. Evaluation Metrics**

Model performance was assessed using accuracy, precision, recall, and F1-score. Results were averaged across folds and reported as mean ± standard deviation to capture model robustness and reproducibility.

**IV. RESULTS**

Table 1 reports the fold-wise validation accuracies and aggregated statistics for all three pretrained models. CoAtNet-Tiny achieved the highest mean validation accuracy (**84.01% ± 0.72**), demonstrating superior overall performance and stability across folds. ViT-B/16 followed closely

(**83.59% ± 2.05**), occasionally surpassing CoAtNet in individual folds but showing higher fold-to-fold variability. VGG16 attained a mean of **81.53% ± 1.65**, confirming its reliability as a CNN baseline but lagging behind the transformer-based models in both accuracy and consistency.

The relatively low standard deviation of CoAtNet-Tiny suggests that its hybrid architecture, combining convolutional token embeddings with hierarchical attention, supports consistent generalization. By contrast, ViT-B/16's larger variability reflects sensitivity to initialization and limited data, common challenges for pure transformer models trained on modest-sized datasets.

| Model | Fold 1 | Fold 2 | Fold 3 | Mean | Std |
|---|---|---|---|---|---|
| VGG16 | 80.87 | 80.31 | 83.40 | 81.53 | 1.65 |
| ViT-B/16 | 82.70 | 85.94 | 82.14 | 83.59 | 2.05 |
| CoatNet-Tiny | 83.4 | 84.81 | 83.83 | 84.01 | 0.72 |

**Table 1: Fold-Wise Validation Accuracies (%)**

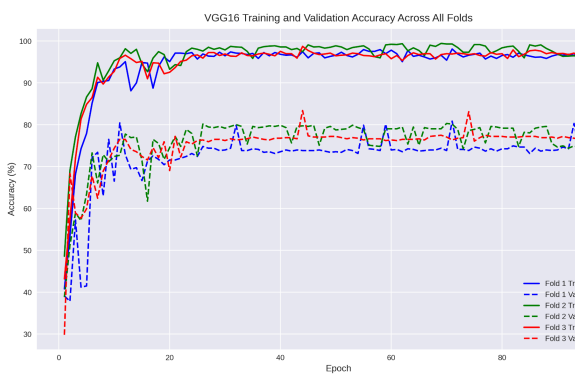

(a)

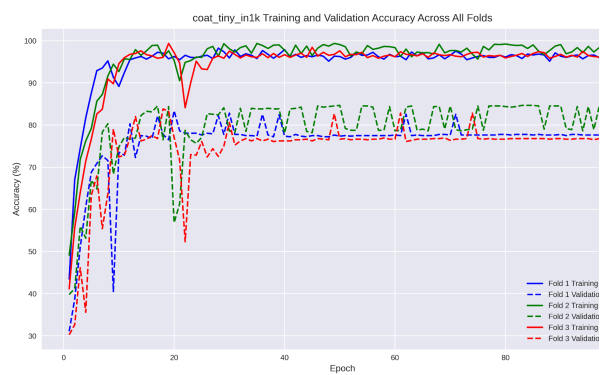

(b)

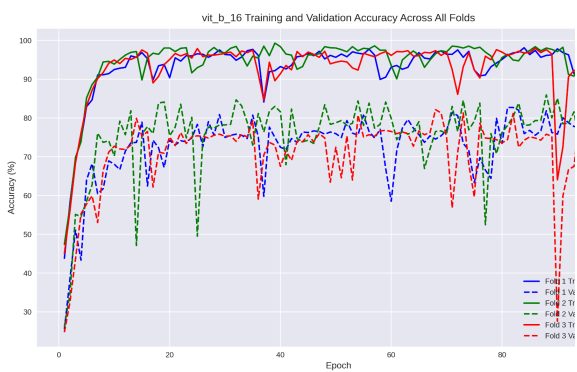

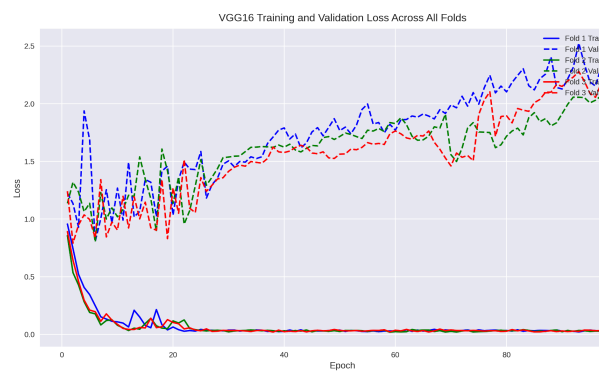

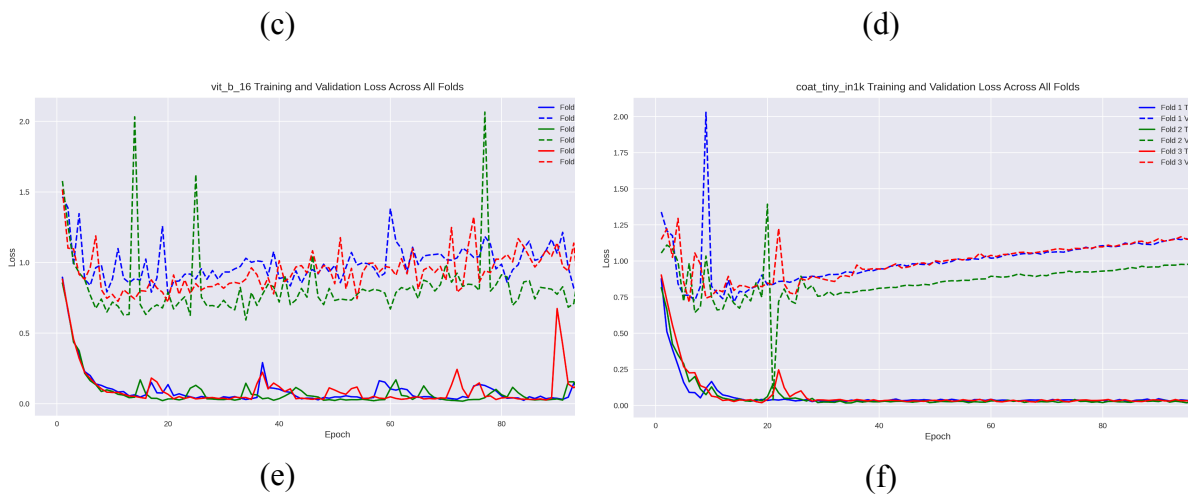

(c)          (d)

(e)          (f)

**Figure 1:** Comparative training and validation performance across all three folds for the three pretrained architectures: (a) training and validation accuracy for VGG16, (b) training and validation accuracy for CoAtNet-Tiny, (c) training and validation accuracy for ViT-B/16, (d) training and validation loss for VGG16, (e) training and validation loss for ViT-B/16, and (f) training and validation loss for CoAtNet-Tiny. Results highlight clear differences in convergence behavior and generalization stability, with CoAtNet-Tiny demonstrating the smoothest learning curves and lowest validation loss across folds.

**A. Training and Validation Dynamics**

Figure 1 presents the training and validation curves for accuracy and loss across all folds. Several key patterns emerge:

**VGG16:**

Training accuracy rose rapidly and plateaued near 97–99%, revealing strong fitting capability. Validation accuracy stabilized around 75–78% with minimal fluctuations, indicating consistency but limited generalization. The training and validation loss curves diverged progressively, training loss approaching zero while validation loss increased beyond 2.0, demonstrating severe overfitting as training progressed.

**ViT-B/16:**

Training accuracy exceeded 95% by approximately Epoch 20, confirming efficient learning. Validation accuracy fluctuated between 70–78%, showing occasional peaks but with noticeable variance across folds. The loss curves revealed pronounced oscillations in validation loss (0.7–1.5) despite stable training loss near zero. These fluctuations point to optimization instability, typical of ViTs trained on limited data. Nevertheless, ViT reached high validation peaks, suggesting strong potential when supported by adequate regularization, warm-up schedules, and data augmentation.

**CoAtNet-Tiny:**

Displayed the most balanced and stable learning pattern among all models. Training accuracy converged to 97–98%, while validation accuracy maintained a consistent 79–82% across folds. Training and validation losses followed smooth, parallel trajectories with only a mild late-epoch increase, signifying excellent generalization and minimal overfitting. CoAtNet's hybrid design, integrating convolutional locality with global attention, appears particularly well suited to the actigraphy-derived image representations used in this study.

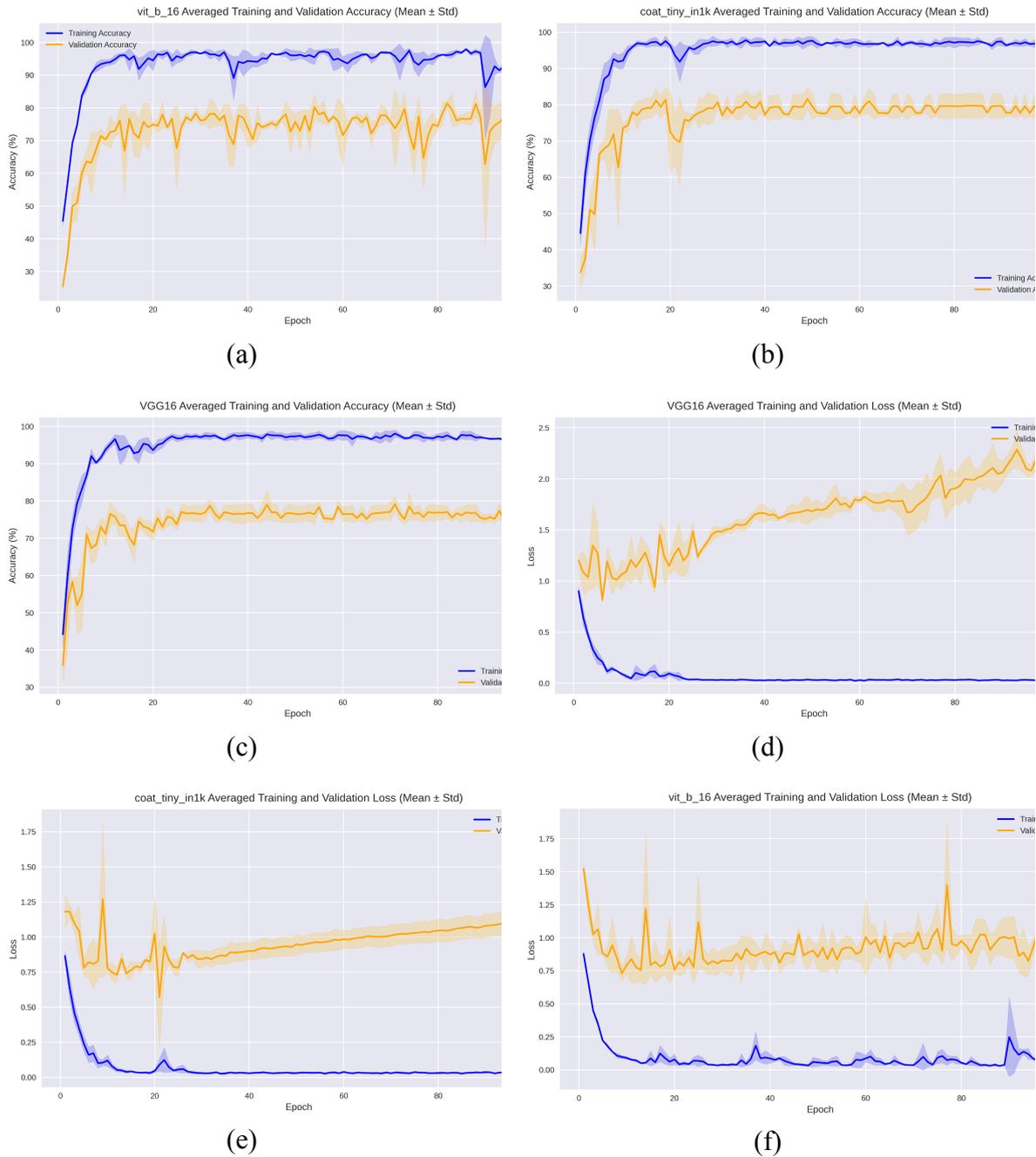

**Figure 2:** Averaged training and validation performance across all folds for the three pretrained architectures showing: (a) averaged training and validation accuracy for VGG16, (b) averaged training and validation accuracy for CoAtNet-Tiny, (c) averaged training and validation accuracy for ViT-B/16, (d) averaged training and validation loss for VGG16, (e) averaged training and validation loss for CoAtNet-Tiny, and (f) averaged training and validation loss for ViT-B/16. Results demonstrate that CoAtNet-Tiny achieved the most stable and lowest validation loss across epochs, ViT-B/16 showed greater fluctuations due to overfitting tendencies, and VGG16 converged more gradually but maintained reliable consistency in both accuracy and loss trajectories.

### B. Averaged Learning Curves

Figure 2 presents the averaged accuracy and loss across all curves (mean ± standard deviation) across folds, offering a clear comparison of convergence stability and generalization behavior.

**Accuracy:**

CoAtNet-Tiny consistently outperformed both ViT-B/16 and VGG16 across most epochs. Its validation curve was smoother and more stable, maintaining high accuracy with narrow variability. ViT-B/16's validation accuracy rose sharply early on but exhibited intermittent dips reflecting instability, while VGG16 trailed with a large, consistent generalization gap despite steady performance.

**Loss:**

CoAtNet-Tiny achieved the lowest and most stable validation loss, decreasing rapidly and then rising only slightly toward the end, indicating mild overfitting. ViT-B/16 displayed persistent divergence between training and validation losses, aligning with its noisier accuracy pattern. VGG16's validation loss increased steadily, confirming progressive overfitting and limited generalization capacity.

Although all three architectures achieved near-perfect training accuracy, only CoAtNet-Tiny effectively balanced fast convergence with stable generalization. Its smaller train–validation gap, minimal standard deviation, and smooth loss trajectory establish it as the most robust and data-efficient model for this actigraphy-based mental-health classification task.

**Detailed Evaluation Metrics:**

Beyond accuracy, Tables 2 and 3 summarize macro- and weighted-averaged precision, recall, and F1-scores.

Table 2: CoAtNet consistently achieved the highest balanced performance (Precision/Recall/F1 ≈ 0.84), reflecting strength across all three classes. ViT and VGG16 trailed slightly, with macro-F1 ≈ 0.81, highlighting their difficulty with minority classes.

Table 3: CoAtNet again led with the best weighted averages (≈0.85), confirming it handled class imbalance better than the others. ViT performed well on the majority class but showed reduced recall on minority conditions, while VGG16 showed reliable but modest scores across classes.

| Model | Precious | Recall | F1-score |
|---|---|---|---|
| VGG16 | 0.81 | 0.80 | 0.80 |
| Vit-B/16 | 0.82 | 0.81 | 0.81 |
| CoatNet-Tiny | 0.84 | 0.84 | 0.84 |

Table 2: Macro-Averaged Metrics (Precision, Recall, F1-score)

| Model | Precision | Recall | F1-score |
|---|---|---|---|
| VGG16 | 0.82 | 0.81 | 0.81 |
| ViT-B/16 | 0.83 | 0.83 | 0.83 |
| CoatNet-Tiny | 0.85 | 0.85 | 0.85 |

Table 3: Weighted-averaged Metrics (Precision, Recall, F1-score)

## V. DISCUSSION

The present study compared convolutional, transformer-based, and hybrid deep learning architectures, VGG16, ViT-B/16, and CoAtNet-Tiny, for the classification of mental health conditions (depression, schizophrenia, and control) from actigraphy-derived images. We hypothesized that hybrid architectures would outperform both traditional convolutional networks and pure transformers by leveraging the strengths of both local feature extraction and global attention mechanisms. Based on our results, we partially confirm this hypothesis: CoAtNet-Tiny consistently achieved superior and more stable performance compared to both VGG16 and ViT-B/16 across folds and evaluation metrics.

The fold-wise accuracies summarized in Table 1 highlight distinct differences in model behavior. CoAtNet-Tiny achieved the highest mean validation accuracy (84.01 % ± 0.72) with minimal variability across folds, indicating stable generalization and robustness. This consistent performance can be attributed to its hybrid design, which integrates convolutional token embedding with hierarchical attention layers. The convolutional front end captures short-term local dependencies inherent in actigraphy-derived patterns, while the transformer component enables modeling of long-range temporal dependencies. Together, these complementary inductive biases appear to align well with the spatio-temporal dynamics of daily activity cycles, yielding strong overall generalization.

By contrast, ViT-B/16 achieved comparable but slightly lower mean validation accuracy (83.59 % ± 2.05) and exhibited larger fold-to-fold fluctuations. Validation curves (see Figure 1) show that ViT reached high training accuracy (>95 % by Epoch 20), but validation performance varied, peaking between 82–86 % before stagnating or declining. This instability reflects ViT's known dependence on large datasets and strong regularization. Given the modest dataset size (~2,133 samples), ViT likely suffered from overfitting, as evident in the widening divergence between training and validation loss curves (see Figure 1). Despite these limitations, ViT's high peak accuracy demonstrates its potential under conditions of abundant data or enhanced regularization techniques such as dropout, Mixup, or stochastic depth.

VGG16, a traditional convolutional architecture, achieved the lowest mean accuracy (81.53 % ± 1.65) but showed consistent and stable training behavior. Its accuracy curves (see Figure 1) display gradual improvement with minimal oscillation, and the validation loss trajectories (see Figure 1) were smooth and monotonic. While its limited representational capacity constrained its performance ceiling, VGG16's resilience and stability reaffirm the reliability of convolutional networks in small-sample biomedical applications.

Figure 2, which depicts averaged training and validation accuracy and loss across folds, further illustrate differences in inductive bias and generalization. CoAtNet-Tiny demonstrated the most balanced learning trajectory, combining fast convergence with stable validation accuracy, indicative of an effective bias–variance trade-off. ViT-B/16 achieved the steepest rise in training accuracy but failed to sustain parallel improvements in validation, reflecting strong overfitting tendencies. VGG16, while slower to converge, maintained steady progress with minimal divergence between training and validation curves, signifying stable optimization.

CoAtNet's validation loss remained the lowest across epochs, suggesting superior generalization and a smoother optimization landscape. ViT's loss gap widened after approximately 30 epochs, consistent with data-scarce transformer overfitting. VGG16's loss plateaued earlier, likely reflecting its inability to capture higher-order temporal relations, though it remained more stable than ViT's.

Further evidence of model differences is provided in Tables 2 and 3, summarizing macro- and weighted-averaged precision, recall, and F1-scores. CoAtNet-Tiny achieved the highest overall macro-F1 (~0.84) and weighted-F1 (~0.85), confirming balanced performance across the three classes. ViT-B/16 reached competitive weighted averages, driven primarily by strong performance on the dominant control class, but displayed reduced macro-F1, reflecting lower sensitivity to minority classes (depression and schizophrenia). VGG16 maintained stable but comparatively lower results across all metrics, performing better on schizophrenia due to its more distinct circadian activity disruptions, but struggling to differentiate depression from control subjects.

These class-specific patterns mirror clinical characteristics of the disorders: depression tends to manifest as subtle alterations in daily activity rhythms, while schizophrenia often results in pronounced circadian dysregulation. Hence, models capable of global pattern integration (like CoAtNet and ViT) performed better on depression, but only CoAtNet achieved consistent generalization across folds.

Taken together, these results indicate that hybrid models such as CoAtNet-Tiny offer the most favorable compromise between convolutional locality and transformer-based global reasoning. CoAtNet's convolutional tokenization ensures that fine-grained local dynamics, such as sleep fragmentation and micro-rest episodes, are effectively encoded, while attention layers capture diurnal context and overall behavioral rhythm. This dual mechanism enables superior generalization from relatively small datasets. Transformer-only architectures like ViT-B/16 remain promising but data-hungry; their performance improves markedly with large-scale pretraining or synthetic augmentation. Classical CNNs such as VGG16 continue to serve as robust baselines for small, homogeneous datasets, though their reliance on local receptive fields limits their discriminative power for complex temporal-spatial dependencies.

Overall, this study provides clear empirical evidence that hybrid convolution-attention architectures outperform both pure CNNs and transformers on actigraphy-based mental health classification.

CoAtNet-Tiny's combination of rapid convergence, low variance, and consistent class-wise accuracy underscores the growing importance of hybrid models in biomedical and behavioral time-series imaging, where limited data and subtle inter-class variations remain enduring challenges.

**C. Limitations and Future Work**

This study has several limitations. The datasets, while representative of actigraphy-based mental health monitoring, are relatively small by modern deep learning standards (~2,133 images). This limited sample size increases the risk of overfitting and restricts external validity, especially for underrepresented conditions such as depression. Class imbalance further constrained generalization, with models tending to favor the majority control class. Additionally, this study focused exclusively on image-based encodings of actigraphy data, omitting multimodal integration with clinical or demographic features that could provide richer contextual information.

Future work should aim to address these limitations through larger, multi-institutional datasets that capture broader demographic and behavioral diversity. Expanding model pipelines to integrate multimodal behavioral, physiological, and clinical data could also enable more comprehensive characterization of mental health states, improving interpretability and robustness in real-world applications.

## VI. CONCLUSION

This study conducted a comparative evaluation of three pretrained deep learning architectures, VGG16 (Simonyan & Zisserman, 2014), ViT-B/16 (Dosovitskiy et al., 2021), and CoAtNet-Tiny (Dai et al., 2021), for the classification of mental health disorders from actigraphy-derived images. Using a subject-wise 3-fold cross-validation setup, CoAtNet-Tiny consistently achieved the best balance between accuracy and generalization, outperforming both classical CNNs and pure transformers. ViT-B/16 demonstrated high peak accuracies but exhibited less stable generalization, reflecting sensitivity to dataset size and overfitting tendencies. VGG16 achieved a lower mean accuracy but remained a reliable and computationally efficient benchmark.

Overall, the results emphasize the promise of hybrid vision architectures for biomedical applications, particularly when dealing with limited or imbalanced datasets. By combining convolutional inductive biases with transformer-based global attention, CoAtNet-Tiny demonstrated robust generalization across both depression and schizophrenia detection tasks. This work extends prior CNN-based approaches by systematically benchmarking multiple pretrained architectures on actigraphy data and highlighting their respective strengths and weaknesses.

Looking forward, the integration of larger datasets, synthetic augmentation pipelines, and multimodal data sources will be key to enhancing clinical applicability. With continued advances in deep learning and wearable sensing technologies, actigraphy-based models could become powerful, low-cost, and non-invasive tools for large-scale mental health screening, supporting early detection and personalized intervention in both clinical and community settings.

## Acknowledgment

ChatGPT (OpenAI) was used solely for language editing. All ideas, analyses, and conclusions are entirely the author's own.